%% file: main.tex
\documentclass[11pt]{article}
\usepackage[preprint]{acl}

\usepackage{times}
\usepackage{latexsym}
\usepackage[T1]{fontenc}
\usepackage[utf8]{inputenc}
\usepackage{inconsolata}

\usepackage{amsmath}
\usepackage{booktabs}
\usepackage{graphicx}
\usepackage{pdflscape}
\usepackage{caption}
\usepackage{xcolor}
\usepackage{framed}
\usepackage{listings}

\title{Do Small Models Use the Law You Give Them?\\
Context-Injected Fine-Tuning for Legal QA in Bangladesh}

\author{
  \textbf{Moniruzzaman Mahadi\textsuperscript{1}},
  \textbf{Abrar Mohammed Tanzim Alam\textsuperscript{1}},
  \textbf{Sayma Siddika Monalisa\textsuperscript{1}}
  \\
  \textbf{Mir Mohammad Asif Abdullah\textsuperscript{1}},
  \textbf{Swakkhar Shatabda\textsuperscript{2}},
  \textbf{Md Adnan Arefeen\textsuperscript{1}}
  \\
  \\[-2pt]
  {\normalfont
    \textsuperscript{1}Department of Electrical and Computer Engineering,
    North South University
  }
  \\
  {\normalfont
    Dhaka, Bangladesh
  }
  \\
  {\normalfont
    \textsuperscript{2}Department of Computer Science and Engineering,
    BRAC University
  }
  \\
  {\normalfont
    Dhaka, Bangladesh
  }
  \\[2pt]
  {\normalfont\small
    \textbf{Correspondence:}
    \href{mailto:momahadi9664@gmail.com}
    {\texttt{momahadi9664@gmail.com}}
  }
}

\begin{document}
\maketitle

\begin{abstract}
A small language model can receive the governing statutory provision and
still answer incorrectly. We test whether fine-tuning on examples containing
relevant law improves later use of retrieved law. We curate 2{,}165 bilingual
QA records from six Bangladeshi acts and three schedules, then fine-tune
Qwen3.5 at 0.8B, 2B, and 4B. Evaluation uses the 2022 and 2023 Bangladesh Bar
Council exams in Bangla and machine-translated English, with no retrieval,
BM25, or FAISS, scored by strict consistency over three seeded runs. At 0.8B,
fine-tuning raises the 2022 English FAISS score from 2 to 34 of 100. Gains at
0.8B and 2B survive paired testing, but the 4B model has no detectable net
gain: Bangla improves while several English conditions regress. Fine-tuning
also reduces answers that drift from Bangla into mostly English from
44.0--53.2\% to 0.2--0.7\%, with adjusted $p<.001$ at every scale. Retrieval
quality is therefore not the only bottleneck. Small bilingual legal models
also differ in how they use supplied law and whether they answer in the
requested language. The dataset is publicly available at
\url{https://huggingface.co/datasets/momahadi/bangladesh-legal-qa-dataset}.
\end{abstract}

\input{sections/introduction}
\input{sections/related-work}
\input{sections/dataset}
\input{sections/methodology}
\input{sections/experiments}
\input{sections/results}
\input{sections/discussion}
\input{sections/conclusion}

\section*{Acknowledgments}

This work was carried out as a Directed Research project at the Department of
Electrical and Computer Engineering, North South University.

\input{sections/limitations}

\bibliography{references}

\clearpage
\appendix
\onecolumn
\input{sections/appendices}

\end{document}

%% file: sections/introduction.tex
\section{Introduction}
\label{sec:intro}

\begin{figure*}[t]
\centering
\includegraphics[width=0.98\textwidth]{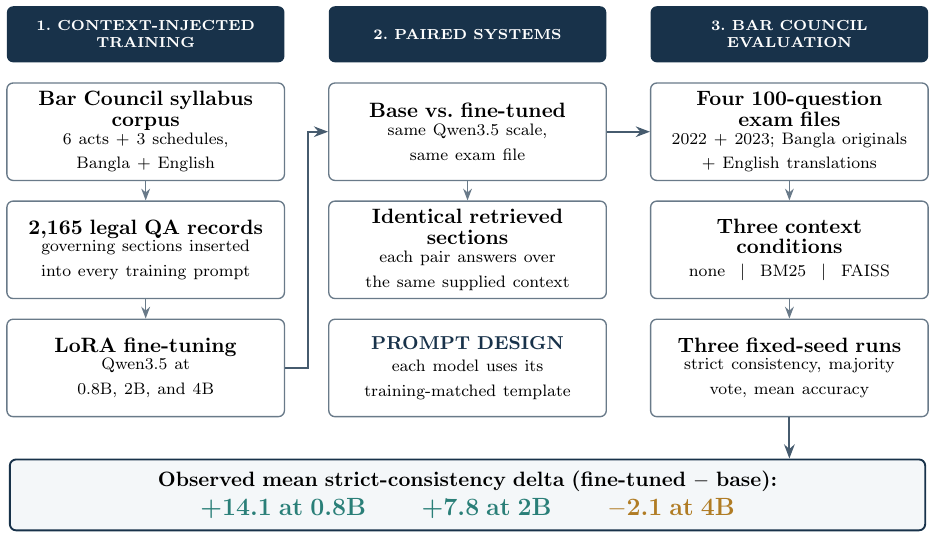}
\caption{Study design. Context-injected supervision produces
fine-tuned variants of Qwen3.5 at three scales. Base and fine-tuned systems
are compared on four Bangladesh Bar Council exam files under no retrieval,
BM25, and FAISS. Mean strict-consistency deltas favor fine-tuning at 0.8B and
2B but reverse at 4B.}
\label{fig:overview}
\end{figure*}

Article 27 of the Constitution of Bangladesh guarantees equality before the
law~\cite{bdconstitution}. Yet legal help remains out of reach for much of the
population. Analyses of the Bangladeshi justice system describe several
million pending cases handled by roughly two thousand judges, prolonged
litigation, unregulated lawyer fees, and narrow eligibility for public legal
aid~\cite{yesmin2025,ahmed2021,jaan2023,tahura2025,islam2024,akter2017}.
Language technology will not fix the court system, but it can lower the
cost of reaching accurate legal information, provided its answers stay
grounded in the governing statute.

Much recent legal NLP pursues that grounding through retrieval: a better
index puts the right provision into the model's context.
Small models strain that assumption. In our experiments, they often failed
with the governing section already in the prompt, and Bangla was worse.
Bengali remains low-resource for current model families~\cite{bhowmik2025}.
Bangladeshi legal language also combines Persian-influenced and colonial-era
terms with standard Bengali~\cite{mina}. Even frontier models have struggled
with authentic Bangladeshi legal queries~\cite{aftahee2025}.

We focus on what happens after retrieval. Does supervised fine-tuning on legal QA
pairs whose prompts already contain the relevant statutory text improve a
small model's use of supplied context at inference time? We call this
ability \emph{context utilization} and measure it as the
change in multiple-choice accuracy with and without retrieved statutory
context, before and after fine-tuning. Figure~\ref{fig:overview} summarizes
the study. Our benchmark is the 2022 and 2023 Bangladesh Bar Council MCQ
examinations, taken in the original Bangla and in machine-translated
English. MINA's best reported cells on these examinations are 75.6--77.0\%,
obtained with the proprietary Gemini-2.5-Flash backend under two-step RAG or
tool use; its small open-weight backends score far lower~\cite{mina}. Those
five-run averages are not directly comparable to our strict-consistency
score. We hold the retrieval machinery fixed and vary the model: base and
fine-tuned Qwen3.5 at 0.8B, 2B, and 4B parameters, each under no retrieval,
BM25, and FAISS dense retrieval, three fixed-seed runs per condition. One
run of GPT OSS 120B gives an informal ceiling; it is not a replicated
baseline.

Our contributions are:
\begin{enumerate}\itemsep1pt
  \item a structured, hierarchy-preserving bilingual corpus of the six
        Bangladeshi acts and three schedules that dominate the Bar Council
        syllabus (Section~\ref{sec:dataset});
  \item a curated bilingual legal QA dataset of 2{,}165 records in three
        difficulty formats, with the statutory context relevant to each
        answer injected into the training prompt
        (Section~\ref{sec:dataset});
  \item a paired comparison of base and context-injected fine-tuned
        models across scales, languages, and retrieval conditions, reported
        under strict-consistency, majority-vote, and mean-accuracy scoring,
        with the regressions made explicit
        (Sections~\ref{sec:setup}--\ref{sec:results}).
\end{enumerate}

The effect depends strongly on model size. The fine-tuned models posted large
gains at 0.8B and 2B, and retrieval helped them much more than their base
counterparts. At 4B, Bangla mostly improved, while several English conditions
regressed. Fine-tuning also removed a large language-drift failure: base
models answered Bangla questions mostly in English in 53.2\%, 44.0\%, and
44.7\% of visible outputs at 0.8B, 2B, and 4B. Fine-tuning reduced those
rates to 0.7\%, 0.2\%, and 0.7\%, respectively, with adjusted $p<.001$ at
every scale. These experiments show gaps in both context utilization and
answer-language control across model scale.

%% file: sections/related-work.tex
\section{Related Work}
\label{sec:related}

\paragraph{Low-resource legal QA for Bangladesh.}
\citet{wasi2024} explored LLM-based legal assistance for Bangladesh by
fine-tuning a GPT-2 variant on an English corpus of Bangladeshi statutes. The
authors describe a proof of concept with substantial gaps in accuracy and
safety. Its English-only setup does not reflect bilingual legal practice.
The same group's MINA~\cite{mina} is a multilingual legal-assistant system
evaluated with several model backends and retrieval configurations on the
2022 and 2023 Bar Council examinations, with law professors reviewing
outputs. Its highest table values, 75.6--77.0\%, are five-run averages from
the proprietary Gemini-2.5-Flash backend under two-step RAG or tool use;
they are not directly comparable to our strict-consistency score. The small
open-weight models in the same evaluation score far lower. MINA evaluates a
complete assistant across backends and retrieves over the full body of
Bangladeshi statute; we instead isolate how small Qwen3.5 systems behave
under controlled retrieval, searching only the six acts and three schedules
of the Bar Council syllabus. MINA also documents
hallucinated provisions, missed statutory conditions, and civil/criminal
confusion, and leaves open whether those failures persist when the relevant
text is supplied correctly; our experiments test exactly that setting.

\paragraph{Retrieval-augmented legal NLP.}
RAG~\cite{rag} grounds generation in retrieved evidence. Over Bangladesh
Police Gazettes, \citet{legalrag} built a bilingual RAG framework with
relevance checking and query refinement before generation.
\citet{legalbenchrag} introduced LegalBench-RAG to evaluate the retrieval
step in isolation. They argue that existing benchmarks take retrieval for
granted and that long or noisy context can bury the governing provision.
LegalBench-RAG evaluates what enters the prompt; our study starts after that
step, fixing retrieval and comparing base against fine-tuned models on
identical retrieved context.

\paragraph{Bengali capability and legal datasets.}
\citet{bhowmik2025} evaluated ten open LLMs across eight translated Bangla
datasets, documenting consistent gaps relative to English, especially for
smaller models and for families whose tokenizers over-fragment Bangla.
\citet{aftahee2025} evaluated four frontier LLMs on 250 authentic questions
from a Bangladeshi legal-advice community with LLM-as-judge scoring plus
licensed lawyers, finding well-structured responses alongside dangerous
misinformation such as fabricated citations; expert review keeps exposing
failures automated scores miss, which also cautions against over-reading
our automated MCQ scores. On the dataset side, IndicLegalQA~\cite{indiclegalqa} evaluated
parameter-efficient fine-tuning on structured Indian legal QA, but did not
test whether training with supplied context improves the \emph{use} of
context at inference time, the distinction central here.

\paragraph{Benchmarks and evaluation validity.}
KoBLEX~\cite{koblex} grounds open legal QA in statutory provisions and asks
for multi-hop reasoning over them; IL-TUR~\cite{iltur} spans a far wider set
of Indian legal understanding and reasoning tasks. SaulLM-7B~\cite{saullm}
relies on continued pretraining over a large English legal corpus and does
not target Bangla statutory QA.

\paragraph{The gap.}
Existing work separately improves legal retrieval, documents how far small
models fall behind in Bangla, and fine-tunes models on legal QA. Within
this reviewed set, we found no evaluation of whether context-injected
supervision helps a small Bangla-capable model use statutory text supplied
at inference time. We test that by fine-tuning small models on bilingual
legal QA whose prompts contain the governing sections, then comparing base
and fine-tuned models on identical retrieved context.

%% file: sections/dataset.tex
\section{Corpus and QA Data}
\label{sec:dataset}

We recomputed every count in this section from the released dataset.
Figure~\ref{fig:pipeline} (Appendix~\ref{app:dataset-detail}) traces the
detailed data, training, and Bar Council evaluation flow.

\subsection{Corpus Scope, Sources, and Structure}
\label{sec:sources}
\label{sec:extraction}

The corpus covers the six acts and three schedules that carry the most
weight in the Bangladesh Bar Council examination syllabus: the Penal Code
1860, Code of Criminal Procedure 1898, Evidence Act 1872, Code of Civil
Procedure 1908, Specific Relief Act 1877, and Limitation Act 1908, plus CPC
Schedule~I (Orders and Rules), CrPC Schedule~II, and Limitation Act
Schedule~I. We chose this scope because the enrolment examination is our
benchmark; the corpus is not the full body of Bangladeshi statute law.
Schedule labels follow the released corpus. Each act is bilingual: English
text from the official Bangladesh laws portal
(\url{http://bdlaws.minlaw.gov.bd/}), Bangla text from BGS Technologies'
published act collections.

Statute documents are irregular: act, part, chapter, section, subsection,
clause, subclause, plus attached illustrations and explanations. We
converted each act to JSON with Google's Gemini on the Antigravity platform,
one custom prompt per act, after regular-expression preprocessing had
separated illustrations from identically labelled clauses. We iterated a
schema-enforcement prompt (Appendix~\ref{app:prompts}) until the files
conformed, then inspected them manually. For every section the corpus keeps
its act, chapter, number, marginal note, full text, and nested structure.

\subsection{QA Generation}
\label{sec:generation}

From the corpus we generated instruction-tuning records in three categories
of increasing difficulty, each with a dedicated prompt per language
(Appendix~\ref{app:prompts}): \textbf{single-hop extraction} (answer from
one supplied provision), \textbf{advanced selection} (five candidate
sections, four semantically similar distractors, governing provision at a
randomized position), and \textbf{bar-exam style} (synthetic scenario MCQs
generated from the statutory corpus, using earlier bar-examination questions
only as style exemplars, with each item annotated with the governing provision,
candidates, and correct option). Generation kept failing in familiar ways:
paraphrased statutory text, skipped clauses, repetitive questions, cross-act
contamination, format drift. We engineered these out through prompt
revision, and the released prompts carry the countermeasures.

\subsection{Quality Control and Composition}
\label{sec:composition}
\label{sec:quality}

We reviewed the records ourselves: fact-checking answers against the source
law, verifying verbatim quotation, deleting or regenerating defective
entries. No per-record review log survives, so we report this as our
process, not as an audited exhaustive protocol.
Every released record carries a per-record \texttt{quality\_flag}, and none
of the 2{,}165 is flagged. We report only the released count because the
pre-filter generation pool was not preserved.

\begin{table}[t]
\centering
\scriptsize
\setlength{\tabcolsep}{2pt}
\begin{tabular}{lrrrr}
\toprule
\textbf{Act / schedule} & \textbf{S-hop} & \textbf{Adv.} & \textbf{Bar} & \textbf{Tot.}\\
 & EN/BN & EN/BN & EN/BN & \\
\midrule
Penal Code 1860        & 90/127 & 111/84 & 39/39 & 490\\
Evidence Act 1872      & 66/94  & 109/75 & 40/33 & 417\\
CrPC 1898              & 66/104 & 28/108 & 59/40 & 405\\
Limitation Act 1908    & 36/42  & 38/43  & 41/40 & 240\\
Specific Relief 1877   & 17/66  & 6/68   & 40/20 & 217\\
CPC 1908               & 27/60  & 13/29  & 29/39 & 197\\
CPC Sch.\ I            & --     & --     & 39/40 & 79\\
CrPC Sch.\ II          & --     & --     & 40/20 & 60\\
Limitation Sch.\ I     & --     & --     & 20/40 & 60\\
\midrule
\textbf{Total} & 302/493 & 305/407 & 347/311 & \textbf{2{,}165}\\
\bottomrule
\end{tabular}
\caption{Final dataset by act, category (single-hop, advanced selection,
bar-exam), and language (English/Bangla). Recomputed from the released
2{,}165-record dataset; schedule rows listed separately rather than folded
into parent acts.}
\label{tab:composition}
\end{table}

Table~\ref{tab:composition} gives the verified composition: 2{,}165 records,
1{,}211 Bangla and 954 English; 795 single-hop, 712 advanced-selection, 658
bar-exam-style. Only the bar-exam category draws on the schedules, matching
how heavily the enrolment examination tests Orders, Rules, and limitation
periods. Each record carries the question, verbatim provision text, candidate
sections where applicable, answer, citation, type, difficulty, language, and
source metadata. How the records are split and trained on is described in
Section~\ref{sec:method}.

\subsection{SFT Packaging: Context Injection}
\label{sec:packaging}

We prepared two chat-format JSONL variants of the 2{,}165 records: a
direct-answer variant, and an IRAC variant that wrote out the reasoning
before the answer. All fine-tuning and evaluation in this paper use the
direct-answer variant. The IRAC variant made each training example much
longer, and fine-tuning on those longer inputs did not run to completion on
our Kaggle setup, so we set it aside. Context injection is the load-bearing
choice. Every training
example looks exactly like a retrieval-augmented inference prompt: the
system message ``You are a Bangladeshi legal assistant. Answer the legal
question using the provided law sections.'' and a ``Retrieved Sections:''
block in the user turn. During training the block is filled deterministically
from the record, governing section plus distractors; no retriever runs. The
model should learn to ground its answers in supplied sections, in the same
format a real retriever will fill later.

\subsection{Evaluation Benchmark}
\label{sec:leakage}

The benchmark is four 100-question files: the 2022 and 2023 Bar Council MCQ
examinations in Bangla, and English versions translated from the originals
with Claude Sonnet 4.5 (not independently re-verified by legal translators).
The training and evaluation sets stay separate by construction. We wrote the
training questions from the Bar Council syllabus statutes, not from the exam
files, and we withheld the 2022 and 2023 exams from generation; the
bar-exam-style prompts used earlier examinations only as format and style
exemplars. We confirmed the separation directly: after Unicode
normalization, whitespace collapsing, and punctuation stripping, none of the
400 evaluation questions matches any of the 2{,}165 training questions
exactly. The only thing the two sets share is the underlying law. During
retrieval-based evaluation, BM25 and FAISS search those same statutes to
supply relevant provisions, which is the intended setup and not exposure to
the benchmark questions or answer keys. Released artifacts are mapped in
Appendix~\ref{app:repro}.

%% file: sections/methodology.tex
\section{Method}
\label{sec:method}

\paragraph{Models.}
We use the Qwen3.5 series at 0.8B, 2B, and 4B parameters~\cite{qwen}, chosen
because its Bangla output stayed stable in preliminary use without decoding
interventions. Training all three scales under one configuration lets model
size vary while data, adapter design, and optimization stay fixed. In
closed-book trials these models answered from parametric memory and
repeatedly reached for Indian rather than Bangladeshi law, citing the Indian
CrPC 1973 in place of the Bangladeshi CrPC 1898. That failure motivated the
grounding-first design.

\paragraph{Training-time injection vs.\ inference-time retrieval.}
Two separate mechanisms put law into the prompt. At
training time, the ``Retrieved Sections'' block is filled deterministically
from the dataset (Section~\ref{sec:packaging}); no retriever is involved. At
inference time, the same block is filled three ways as experimental
conditions: left empty, filled by BM25, or filled by FAISS dense retrieval.
The hypothesis is that a model fine-tuned on the deterministic version
extracts more value from the retrieved version than its base counterpart.

\paragraph{Fine-tuning.}
We trained LoRA~\cite{lora} adapters via Unsloth~\cite{unsloth}
($r{=}32$, $\alpha{=}64$, dropout 0.05) on all seven linear projections;
about 1\% of parameters were trainable (12.8M of
0.87B at 0.8B, 21.8M of 2.2B at 2B, 42.5M of 4.6B at 4B). We requested
QLoRA-style~\cite{qlora} 4-bit base-model loading in every training run, but
we cannot confirm that the quantization engaged for this architecture: our
logs record only a compute-dtype fallback to float32, and we kept no
per-layer quantization check or memory figure that would settle it. We
therefore report the configuration we set and stop short of claiming whether
the base weights ran in 4-bit or full precision during training. Training used AdamW
8-bit, learning rate $2{\times}10^{-4}$ with cosine decay, weight decay
0.01, 2 epochs, and 2{,}048-token sequences, with loss on assistant tokens
only and thinking mode disabled (full configuration in
Appendix~\ref{app:training}). In practice we trained with an effective
batch of 32 over 122 optimizer steps at every scale, and 14 warmup steps,
about 11\% of the schedule. We had configured a batch of 16 and a 6\% warmup,
so these are the values the runs actually used, and we report the executed
numbers throughout. Our 90/10 split
(seed 42) gives 1{,}948 training and 217 validation examples with no
over-length exclusions, the same at 0.8B, 2B, and 4B.
We reused the fine-tuning system prompt verbatim at inference, keeping the
training and inference templates matched; fine-tuned models can be sensitive
to template changes between the two stages~\cite{lyu2024}. After training,
we merged each adapter into the bfloat16 base-model weights and exported the
merged model to F16 GGUF for evaluation; this merge and export precision is
separate from the training precision.

%% file: sections/experiments.tex
\begin{table*}[t]
\centering
\small
\setlength{\tabcolsep}{3.4pt}
\begin{tabular}{l *{12}{c}}
\toprule
& \multicolumn{6}{c}{\textbf{2022}} & \multicolumn{6}{c}{\textbf{2023}}\\
\cmidrule(lr){2-7}\cmidrule(lr){8-13}
& \multicolumn{3}{c}{\textbf{English}} & \multicolumn{3}{c}{\textbf{Bangla}} & \multicolumn{3}{c}{\textbf{English}} & \multicolumn{3}{c}{\textbf{Bangla}}\\
\cmidrule(lr){2-4}\cmidrule(lr){5-7}\cmidrule(lr){8-10}\cmidrule(lr){11-13}
\textbf{Model} & None & BM25 & FAISS & None & BM25 & FAISS & None & BM25 & FAISS & None & BM25 & FAISS\\
\midrule
GPT OSS 120B$^\dagger$ & 59&67&72&36&43&42&42&58&57&40&43&41\\
\midrule
Qwen3.5 0.8B & 7&4&2&0&3&1&3&0&0&0&0&2\\
Qwen3.5 0.8B FT & \textbf{22}&\textbf{25}&\textbf{34}&\textbf{2}&\textbf{10}&\textbf{8}&\textbf{10}&\textbf{25}&\textbf{21}&\textbf{8}&\textbf{19}&\textbf{7}\\
\addlinespace
Qwen3.5 2B & \textbf{18}&16&18&4&4&6&\textbf{13}&16&17&3&2&3\\
Qwen3.5 2B FT & \textbf{18}&\textbf{32}&\textbf{26}&\textbf{13}&\textbf{13}&\textbf{12}&11&\textbf{24}&\textbf{28}&\textbf{9}&\textbf{19}&\textbf{9}\\
\addlinespace
Qwen3.5 4B & \textbf{32}&\textbf{44}&\textbf{46}&15&16&\textbf{14}&\textbf{29}&\textbf{40}&\textbf{38}&\textbf{14}&23&12\\
Qwen3.5 4B FT & 23&36&45&\textbf{16}&\textbf{21}&10&19&37&34&\textbf{14}&\textbf{24}&\textbf{19}\\
\bottomrule
\end{tabular}
\caption{Strict consistency, the primary score: a question counts as correct
only when all three fixed-seed runs answered it correctly, out of 100
questions per cell. Within each base/fine-tuned pair the better value per
column is bold (ties both bold). $^\dagger$GPT OSS 120B was run once as an
informal reference, so repeat scoring does not apply and its row repeats the
single-run accuracies.}
\label{tab:strict}
\end{table*}

\section{Experimental Setup}
\label{sec:setup}

\paragraph{Systems and runtime.}
Six local systems (base and fine-tuned Qwen3.5 at 0.8B, 2B, 4B) ran as GGUF
models via llama-cpp-python with an 8{,}192-token context on Kaggle dual-T4
instances. Base models ran BF16 GGUF conversions of the official checkpoints, and
fine-tuned models ran merged F16 exports, the same at all three scales. GPT OSS 120B
(\texttt{openai/gpt-oss-120b})~\cite{gptoss} was queried once per condition
through the Groq API at low reasoning effort as a single-run informal
ceiling, not a replicated baseline.

\paragraph{Conditions.}
The three conditions differ only in how the ``Retrieved Sections'' block is
filled: empty (\emph{none}); top-5 sections by BM25~\cite{bm25} over the
deduplicated section-level corpus of Section~\ref{sec:sources}; or top-5 by
exact inner-product search (FAISS \texttt{IndexFlatIP})~\cite{faiss} over
\texttt{paraphrase-multilingual-MiniLM-L12-v2} embeddings~\cite{minilm}.
Retrieved context was capped at 1{,}500 tokens. Both retrievers are
deterministic, so across repeated runs the only stochastic component is
generation.

\paragraph{Decoding and repetition.}
All models used temperature 0.7, top-p 0.8, top-k 20, repetition penalty
1.05, and 512 max output tokens. Every condition ran three times with seeds 42,
456, and 2024, applied identically across variants, modes, and exam files.

\paragraph{Prompt design and answer contract.}
\label{sec:prompt-design}
The fine-tuned models reused their training template, whose trained answer
convention states the option letter together with the option text
(``Therefore, (C) Property only is correct.''). The pairing matters because
legal prose is full of parenthesized labels. ``Section 12(A)'' contains a
perfectly good option letter, and a parser that accepts bare letters will
sometimes score the citation, not the answer; only a letter sitting next to
its option text is safe to trust. In piloting, base models given the
training template did not pick up this convention, so they received an
instruction-following prompt that demands the same contract in words: be
concise, avoid \texttt{<think>} tags, analyse briefly, and end with exactly
one line of the form ``Answer: (X) full option text'', with nothing after
it. Questions, retrieved sections, decoding parameters, and seeds stayed
identical within each base and fine-tuned pair.

\paragraph{Parsing and metrics.}
A rule-based parser extracts the chosen option letter; unparseable outputs
score incorrect, and we keep them with null predictions. We had made the
stored predictions with a permissive extractor, so we later went back
and checked every one of the 21,600 stored outputs against the
letter-plus-text contract. That check confirms 83.5\% of the predictions
through an adjacent letter-text pair, finds 5.4\% unparseable either way,
and turns up conflicting letters in 2.4\%. We adjudicated a sample of those
conflicts by hand and found no systematic winner between the two readings;
re-parsing everything the other way changes no qualitative finding
(Appendix~\ref{app:error-prevalence}). Three scores are
reported separately per model, year, language, and condition over the same
three runs: \textbf{strict consistency} (all three runs correct),
\textbf{majority vote} (at least two correct), and \textbf{mean accuracy}
(mean $\pm$ sample SD, $n{=}3$), each out of 100 questions. We recomputed
every table from the per-run, per-question result files, and they matched our
stored score report in every cell. For the primary strict-consistency outcome, we
compared the paired binary result for each question within every scale,
exam-language, and retrieval cell using a two-sided exact McNemar test. Holm
correction controlled family-wise error across all 36 tests.

We also tested the overall effect at each scale using every stored output.
A two-sided exact paired randomization test swapped the base and fine-tuned
labels once per original exam question, keeping its English and Bangla forms,
three retrieval conditions, and three seeded runs in the same cluster. This
gave 200 question clusters per scale, containing 1,200 strict
question-condition outcomes and 3,600 individual answers for each model
variant. Holm correction was applied separately across the three scales for
the strict and answer-level families. GPT OSS 120B was excluded because it has
one run and no matched fine-tuned variant. Majority-vote comparisons remain
descriptive.

%% file: sections/results.tex
\section{Results}
\label{sec:results}

Table~\ref{tab:strict} reports the primary strict-consistency grid;
Table~\ref{tab:summary} aggregates it across the twelve matched year,
language, and retrieval cells for each model variant. The majority-vote and
mean-accuracy tables appear in Appendix~\ref{app:full-results}, where
Figure~\ref{fig:deltas} also plots all 36 strict-score deltas.

\begin{table}[t]
\centering
\scriptsize
\setlength{\tabcolsep}{2pt}
\begin{tabular}{l rrr r c}
\toprule
& \multicolumn{3}{c}{\textbf{Mean strict accuracy}} & \textbf{Retrieval} & \textbf{FT vs.\ base}\\
\cmidrule(lr){2-4}
\textbf{Model} & Overall & English & Bangla & \textbf{gain} & FT W--T--L\\
\midrule
0.8B base & 1.8 & 2.7 & 1.0 & -1.0 & -- \\
0.8B FT & \textbf{15.9} & \textbf{22.8} & \textbf{9.0} & \textbf{+8.1} & 12/0/0 \\
\addlinespace
2B base   & 10.0 & 16.3 & 3.7 & +0.8 & -- \\
2B FT & \textbf{17.8} & \textbf{23.2} & \textbf{12.5} & \textbf{+7.6} & 10/1/1 \\
\addlinespace
4B base   & \textbf{26.9} & \textbf{38.2} & 15.7 & +6.6 & -- \\
4B FT & 24.8 & 32.3 & \textbf{17.3} & \textbf{+10.2} & 4/1/7 \\
\bottomrule
\end{tabular}
\caption{Aggregates of the strict grid (Table~\ref{tab:strict}); FT =
fine-tuned. Mean strict accuracy averages the twelve
year-language-retrieval cells (Overall), the six English cells, and the six
Bangla cells. Retrieval gain is a model's average BM25 and FAISS cells
minus its no-retrieval cells. FT W--T--L counts the fine-tuned model's wins,
ties, and losses against its base over the twelve cells; the better value
in each pair is bold.}
\label{tab:summary}
\end{table}

\paragraph{Small models benefit most from fine-tuning.}
The 0.8B model gains in all twelve strict cells. Its cross-cell mean rises
from 1.8 to 15.9, and the single largest jump, from 2 to 34 correct
answers, comes on 2022 English with dense retrieval. The 2B story is nearly
as one-sided: ten cells improve, one ties, one regresses, and the mean
climbs from 10.0 to 17.8. Majority vote and mean accuracy move the same way
(Appendix~\ref{app:full-results}), so the gains are not just a stability
effect.

\paragraph{Fine-tuned models make more of the same retrieval.}
Retrieval barely moves the base models at these scales. Averaged over BM25
and FAISS, it shifts the 0.8B base model by -1.0 strict points relative to
no retrieval; the same retrieved sections shift its fine-tuned counterpart
by +8.1. At 2B the split is +0.8 against +7.6. Each pair reads identical
context, so the gap cannot come from retrieval quality. The fine-tuned
systems do more with the same supplied text.

\paragraph{Fine-tuning does not lift overall 4B accuracy.}
At 4B the pattern breaks. Fine-tuning lowers the cross-cell mean by 2.1
points and loses seven of twelve strict comparisons. The damage sits in
English, where the mean falls from 38.2 to 32.3; Bangla still rises, from
15.7 to 17.3. This base model already turns retrieval into a +6.6 gain, so
there was less room to improve, and the 4B comparison shows no consistent
benefit from fine-tuning.

\paragraph{Every system scores higher in English.}
The gap holds in every matched condition, at every scale. Even the
single-run GPT OSS 120B reference reaches 72 on 2022 English with dense
retrieval but only 42 on the corresponding Bangla exam; that row is an
informal ceiling, not a replicated baseline.

\paragraph{Fine-tuning nearly eliminates language drift.}
Language adherence changes more uniformly than accuracy. On Bangla exams, the
base models answer mostly in English in 44.0\% to 53.2\% of visible outputs.
After fine-tuning, every rate falls below 1\%
(Table~\ref{tab:language-drift}). All three reductions survive
question-clustered paired testing after Holm correction.
Appendix~\ref{app:language-drift} gives the exact counts and full protocol.

\begin{center}
\scriptsize
\setlength{\tabcolsep}{5pt}
\begin{tabular}{lrrrr}
\toprule
\textbf{Scale} & \textbf{Base (\%)} & \textbf{FT (\%)} & $\boldsymbol{\Delta}$ \textbf{points} & $\boldsymbol{p_{\mathrm{adj}}}$ \\
\midrule
0.8B & 53.2 & 0.7 & -52.5 & $<.001$ \\
2B   & 44.0 & 0.2 & -43.7 & $<.001$ \\
4B   & 44.7 & 0.7 & -44.0 & $<.001$ \\
\bottomrule
\end{tabular}
\captionsetup{hypcap=false}
\captionof{table}{Language drift on Bangla exams. Each value is the share of
visible outputs written mostly in English. Negative changes favor the
fine-tuned model. Tests preserve the nine outputs attached to each original
question as one cluster.}
\label{tab:language-drift}
\end{center}

\paragraph{Paired tests support gains at 0.8B and 2B, but not 4B.}
The gains at the two smaller scales survive paired testing; the 4B change
does not. The scale-level test keeps translations, retrieval modes, and
repeated generations for each of 200 underlying questions in the same
cluster. Table~\ref{tab:scale-significance} shows significant gains on both
outcomes at 0.8B and 2B, but neither outcome reaches significance at 4B.
Appendix~\ref{app:significance} reports the complementary 36 cell-level
McNemar tests.

\begin{center}
\scriptsize
\setlength{\tabcolsep}{4pt}
\begin{tabular}{l*{2}{r}@{\hskip 14pt}*{2}{r}}
\toprule
& \multicolumn{2}{c}{\textbf{Strict consistency}} & \multicolumn{2}{c@{}}{\textbf{All answers}} \\
\cmidrule(lr){2-3}\cmidrule(l){4-5}
\textbf{Scale} & \multicolumn{1}{c}{$\Delta$/1,200} & \multicolumn{1}{c}{$p_{\mathrm{adj}}$} & \multicolumn{1}{c}{$\Delta$ points} & \multicolumn{1}{c@{}}{$p_{\mathrm{adj}}$} \\
\midrule
0.8B & +169 (+14.1) & $<.001$ & +7.25 & $<.001$ \\
2B   & +94 (+7.8)   & $<.001$ & +6.28 & $<.001$ \\
4B   & -25 (-2.1)   & .266    & -0.78 & .596 \\
\bottomrule
\end{tabular}
\captionsetup{hypcap=false}
\captionof{table}{Scale-level paired randomization tests clustered by the
200 original exam questions per scale (100 from 2022 and 100 from 2023).
``All answers'' uses all 3,600 binary outputs
per variant and scale. Holm adjustment is across the three scales separately
for each outcome family.}
\label{tab:scale-significance}
\end{center}

%% file: sections/discussion.tex
\section{Discussion}
\label{sec:discussion}

Context-injected fine-tuning helps most where the base model starts
weakest. At 0.8B and 2B, base models gain little from retrieval; their
fine-tuned counterparts gain 7.6 to 8.1 strict points, and the paired tests
of Table~\ref{tab:scale-significance} confirm the difference is not noise.
If retrieval quality alone set the ceiling, both members of a pair should
have benefited similarly from identical retrieved text. They did not. At
4B the base model already turns retrieval into a 6.6-point gain, fine-tuning
adds no overall accuracy benefit, and the change is not statistically
distinguishable from zero. Context injection appears most useful when the
base model has not already picked up that behavior on its own.

These experiments measure output behavior, not mechanism. A correct option
does not prove that a model used the supplied statute, and strict
consistency mixes correctness with stability. Majority vote and mean
accuracy address the stability part; reliance on the statute stays
unverified either way.

Several factors could explain the 4B English regressions. The leading suspect
is the training recipe: LoRA settings tuned at 0.8B were reused at 4B, and
warmup reached 11\% rather than the planned 6\%. Thinking-mode leakage is the
measurable alternative. Across 36 cells, higher leakage accompanies smaller
gains (Spearman $\rho=-0.47$; Table~\ref{tab:leakcells}), and answers trapped
inside think blocks usually score as wrong. Short direct-answer supervision
may also displace useful behavior where the base model is strongest, which
fits the English decline and Bangla improvement. These runs cannot isolate
the causes.

The language result is less ambiguous than the accuracy pattern. Fine-tuning
drives majority-English replies to Bangla questions below 1\% at all three
scales (Table~\ref{tab:language-drift}). This measures response-language
adherence, not legal correctness or fine-grained code-switching. The output
audit also finds fewer parse failures but new thinking-mode leakage at 2B and
4B (Appendix~\ref{app:error-prevalence}). In two traceable cases, a correct
option rests on faulty reasoning, while a wrong option includes fluent but
fabricated law
(Appendix~\ref{app:errors}). None of these diagnostics estimates legal
reliability.

%% file: sections/conclusion.tex
\section{Conclusion}
\label{sec:conclusion}

Context-injected fine-tuning produced clear gains at 0.8B and 2B on the two
bilingual Bangladesh Bar Council exams, and both survive paired significance
testing. At those scales the fine-tuned models also did much more with the
same retrieved statutory context than their base counterparts. Across all
three scales, it reduced majority-English replies to Bangla questions from
rates between 44.0\% and 53.2\% to rates below 1\%. The effect on answer
language is consistent even though the 4B accuracy result is not. At 4B,
Bangla accuracy improves modestly while several English conditions regress,
leaving no detectable net
effect. The released corpus, 2{,}165-record dataset, and per-question outputs
make the comparison auditable and reproducible.

%% file: sections/limitations.tex
\section*{Limitations}
\label{sec:limitations}

The GGUF formats differ slightly, and the 4B base evaluation ran in a Kaggle
notebook environment that was not itself preserved, though its notebook
source and per-question outputs were.
The LoRA and optimization configuration was selected for the 0.8B model and
reused unchanged at 2B and 4B; we suspect the larger models needed
scale-specific settings, and that reusing this recipe contributes to the
4B English regressions. The effective batch (32) and warmup share
(11\%) we ran also differ from the targets we configured (16 and 6\%), as
reported in Section~\ref{sec:method}. The permissive answer parser can confuse
option letters in citations or restated options with the final choice; our
letter-plus-text audit of all stored outputs bounds such conflicts at
2.4\%, with no systematic direction under adjudication
(Appendix~\ref{app:error-prevalence}).

Multiple-choice accuracy is a narrow proxy for legal QA, while strict
consistency also rewards cross-seed stability. Cell-level McNemar tests and
scale-level question-clustered randomization tests assess the primary strict
outcome; the latter also tests all individual answers without assuming that
translations, retrieval modes, or seeded repetitions are independent. These
tests were added after inspection of the results and were not preregistered.
The two exam years and 200 underlying questions still limit power and
generalization; majority-vote comparisons remain descriptive. The English sets are machine translations
without independent legal review. The 2022 and 2023 benchmark questions were
withheld from synthetic-data generation, and the exact-match audit found no
evaluation question copied into training. Possible exposure of the pretrained
base checkpoints to these public examinations is outside our data pipeline and
was not auditable. Repeated development use of the same 400 questions also
creates benchmark co-adaptation risk.

The study covers six acts, three schedules, two exam years, one model family,
and MCQs only. It does not establish performance across Bangladeshi law or on
open-ended legal work. The corpus and QA data were partly LLM-generated; the
authors reviewed them, but no per-item review record supports a claim of
exhaustive legal verification. Preserved and missing artifacts are listed in
Appendix~\ref{app:repro}.

\section*{Ethics Statement}
These systems are research artifacts, not legal-advice tools. Even the best
local model answers most Bangla questions incorrectly under strict scoring,
and inspected outputs include confident fabricated citations. Deployment
would require qualified human oversight, provenance checks, and broader
evaluation. The released datasets contain public statutory and exam text and
no personal data.

%% file: sections/appendices.tex
\section{Dataset and Training Details}
\label{app:dataset-detail}
\label{app:training}

\begin{figure}[ht]
\centering
\includegraphics[width=.92\textwidth]{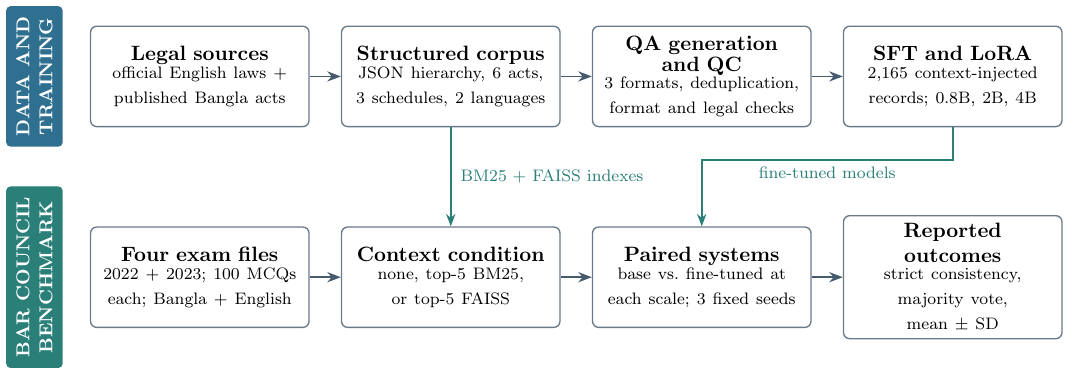}
\caption{Detailed artifact flow. The upper row builds context-injected
fine-tuning data and models; the lower row evaluates base and fine-tuned
systems on four Bangladesh Bar Council exam files. The structured corpus also
supplies the BM25 and FAISS retrieval indexes.}
\label{fig:pipeline}
\end{figure}

\begin{table}[ht]
\centering
\small
\begin{tabular}{ll}
\toprule
\textbf{Component} & \textbf{Setting} \\
\midrule
Training method & LoRA via Unsloth; 4-bit loading requested, not confirmed by logs \\
LoRA & $r=32$, $\alpha=64$, dropout 0.05; seven linear projections \\
Optimizer & AdamW 8-bit, learning rate $2\times10^{-4}$, cosine decay \\
Schedule & 2 epochs, weight decay 0.01, cosine warmup (14 steps) \\
Batch and sequence & effective batch 32 (executed; config targeted 16); 2,048 tokens \\
Split & 1,948 train / 217 validation, seed 42 \\
Inference & temperature 0.7, top-p 0.8, top-k 20 \\
Generation & repetition penalty 1.05; 512 output tokens \\
Runs & seeds 42, 456, 2024 \\
Retrieval & top-5 BM25 or FAISS; 1,500-token context cap \\
\bottomrule
\end{tabular}
\caption{Training and inference configuration.}
\label{tab:training}
\end{table}

\clearpage
\section{Full Results}
\label{app:full-results}

Table~\ref{tab:strict} in the main text reports the primary
strict-consistency grid. Tables~\ref{tab:majority} and~\ref{tab:meansd}
complete it with majority-vote and mean-accuracy scoring over the same runs.

\begin{table}[ht]
\centering
\scriptsize
\setlength{\tabcolsep}{4pt}
\begin{tabular}{l *{12}{c}}
\toprule
& \multicolumn{6}{c}{\textbf{2022}} & \multicolumn{6}{c}{\textbf{2023}}\\
\cmidrule(lr){2-7}\cmidrule(lr){8-13}
& \multicolumn{3}{c}{\textbf{English}} & \multicolumn{3}{c}{\textbf{Bangla}} & \multicolumn{3}{c}{\textbf{English}} & \multicolumn{3}{c}{\textbf{Bangla}}\\
\cmidrule(lr){2-4}\cmidrule(lr){5-7}\cmidrule(lr){8-10}\cmidrule(lr){11-13}
\textbf{Model} & None & BM25 & FAISS & None & BM25 & FAISS & None & BM25 & FAISS & None & BM25 & FAISS\\
\midrule
GPT OSS 120B$^\dagger$ & 59&67&72&36&43&42&42&58&57&40&43&41\\
\midrule
Qwen3.5 0.8B & 35&30&32&22&18&17&23&24&26&12&22&14\\
Qwen3.5 0.8B FT & 40&45&53&18&24&25&27&39&39&27&33&23\\
\addlinespace
Qwen3.5 2B & 39&47&42&20&20&11&28&36&34&14&18&13\\
Qwen3.5 2B FT & 43&50&52&28&31&32&28&37&38&25&27&25\\
\addlinespace
Qwen3.5 4B & 47&58&59&27&36&28&47&53&50&38&37&32\\
Qwen3.5 4B FT & 45&59&63&35&37&30&42&52&53&36&43&38\\
\bottomrule
\end{tabular}
\caption{Majority vote: at least two of three runs are correct. Each cell
contains 100 questions. $^\dagger$Single run.}
\label{tab:majority}
\end{table}

\begin{table}[ht]
\centering
\scriptsize
\setlength{\tabcolsep}{1.2pt}
\begin{tabular}{l *{12}{c}}
\toprule
& \multicolumn{6}{c}{\textbf{2022}} & \multicolumn{6}{c}{\textbf{2023}}\\
\cmidrule(lr){2-7}\cmidrule(lr){8-13}
& \multicolumn{3}{c}{\textbf{English}} & \multicolumn{3}{c}{\textbf{Bangla}} & \multicolumn{3}{c}{\textbf{English}} & \multicolumn{3}{c}{\textbf{Bangla}}\\
\cmidrule(lr){2-4}\cmidrule(lr){5-7}\cmidrule(lr){8-10}\cmidrule(lr){11-13}
\textbf{Model} & None & BM25 & FAISS & None & BM25 & FAISS & None & BM25 & FAISS & None & BM25 & FAISS\\
\midrule
0.8B & 36.0$\pm$8.5&37.0$\pm$9.5&34.7$\pm$11.0&23.7$\pm$9.5&23.0$\pm$5.0&22.0$\pm$10.6&28.7$\pm$1.2&29.7$\pm$13.8&30.0$\pm$12.1&21.3$\pm$2.3&22.3$\pm$10.7&22.7$\pm$6.4\\
0.8B FT &42.3$\pm$3.1&43.7$\pm$2.3&52.7$\pm$2.3&23.0$\pm$5.0&25.7$\pm$1.5&29.0$\pm$6.2&30.3$\pm$1.5&41.3$\pm$3.5&38.7$\pm$1.2&30.0$\pm$6.1&33.3$\pm$1.2&28.0$\pm$3.0\\
\addlinespace
2B &41.3$\pm$0.6&43.0$\pm$5.3&43.3$\pm$4.5&25.7$\pm$2.1&27.0$\pm$4.4&21.0$\pm$7.0&33.3$\pm$2.3&38.3$\pm$3.1&35.7$\pm$4.0&19.3$\pm$3.5&24.0$\pm$3.5&18.3$\pm$4.7\\
2B FT &45.7$\pm$4.0&50.7$\pm$1.5&47.7$\pm$4.7&33.3$\pm$3.8&34.0$\pm$5.2&31.7$\pm$2.3&34.0$\pm$4.6&39.7$\pm$2.1&40.7$\pm$4.5&26.7$\pm$3.8&32.0$\pm$0.0&29.7$\pm$6.1\\
\addlinespace
4B &51.7$\pm$3.5&58.7$\pm$3.2&61.3$\pm$4.0&35.7$\pm$6.7&40.3$\pm$1.5&33.0$\pm$2.6&50.0$\pm$2.6&56.0$\pm$2.6&53.7$\pm$3.2&39.0$\pm$4.0&41.7$\pm$4.5&34.3$\pm$1.5\\
4B FT &47.0$\pm$4.6&57.7$\pm$1.5&62.7$\pm$0.6&37.7$\pm$0.6&41.0$\pm$7.0&33.0$\pm$2.0&41.3$\pm$1.2&55.0$\pm$5.6&54.3$\pm$3.5&34.7$\pm$4.6&43.0$\pm$2.6&38.7$\pm$2.9\\
\bottomrule
\end{tabular}
\caption{Mean per-run accuracy $\pm$ sample standard deviation ($n=3$) for
the six Qwen3.5 systems (model size, base or fine-tuned).}
\label{tab:meansd}
\end{table}

\begin{figure}[ht]
\centering
\includegraphics[width=.92\textwidth]{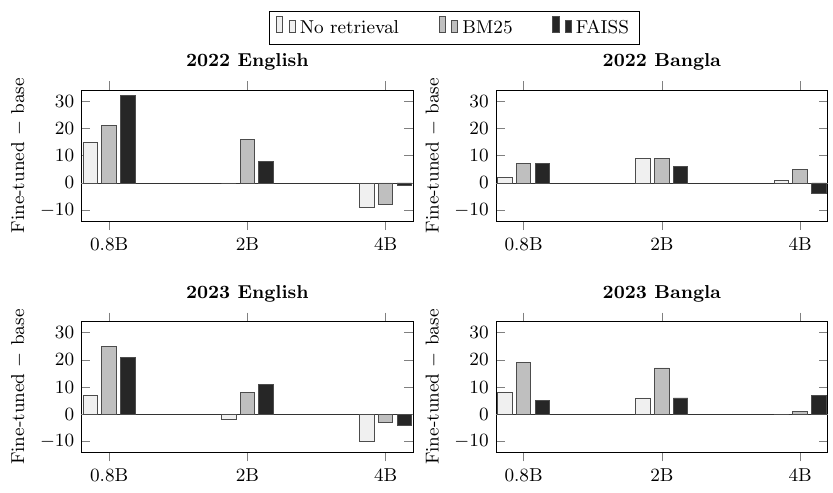}
\caption{Fine-tuned minus base strict-consistency score (out of 100) for
every scale, year, language, and retrieval condition, computed from
Table~\ref{tab:strict}. No retrieval leaves the context block empty; BM25
and FAISS fill it with the top-5 retrieved sections. All panels share the
same y-axis scale; bars below zero are regressions.}
\label{fig:deltas}
\end{figure}

\clearpage
\section{Cell-Level Paired Significance Tests}
\label{app:significance}

\begin{figure}[ht]
\centering
\includegraphics[width=.98\textwidth]{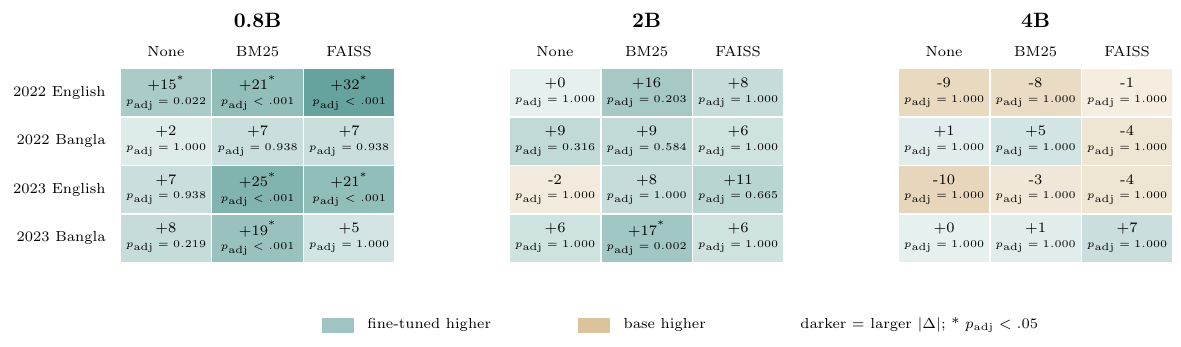}
\caption{Fine-tuned-minus-base strict-consistency difference and
Holm-adjusted $p$ value for each scale, exam language, year, and retrieval
condition. Stars mark adjusted $p<.05$.}
\label{fig:significance}
\end{figure}

Figure~\ref{fig:significance} gives the secondary cell-level analysis behind
the scale-level tests in Table~\ref{tab:scale-significance}. Each McNemar test
uses the same 100 questions for the base and fine-tuned variants. Holm
adjustment controls family-wise error across all 36 cells.

\section{Error Prevalence and Parser Audit}
\label{app:error-prevalence}

Table~\ref{tab:prevalence} counts three failure modes over every stored
output, 1{,}800 per row: two exam years, three retrieval conditions, and
three runs of 100 questions. Parse failure means the stored prediction is
null. Think leakage means the output contains a model-emitted
\texttt{<think>} block. Language drift is computed on Bangla exams over
outputs with visible non-think text: the share whose visible answer is
majority non-Bangla script.

\begin{table}[ht]
\centering
\small
\setlength{\tabcolsep}{4.5pt}
\begin{tabular}{ll@{\hskip 14pt}*{2}{r}@{\hskip 18pt}*{2}{r}@{\hskip 18pt}r}
\toprule
& & \multicolumn{2}{c}{\textbf{Parse fail \%}} & \multicolumn{2}{c}{\textbf{Think leak \%}} & \multicolumn{1}{c}{\textbf{Drift \%}}\\
\cmidrule(lr){3-4}\cmidrule(lr){5-6}\cmidrule(l){7-7}
\textbf{Scale} & \textbf{Var.} & \multicolumn{1}{c}{Bn} & \multicolumn{1}{c}{En} & \multicolumn{1}{c}{Bn} & \multicolumn{1}{c}{En} & \multicolumn{1}{c}{Bn}\\
\midrule
0.8B & base & 19.1 & 8.4 & 0.0 & 0.0 & 53.2\\
0.8B & FT   & 0.8  & 0.2 & 2.9 & 3.7 & 0.7\\
\addlinespace
2B   & base & 26.1 & 16.7 & 0.0 & 0.0 & 44.0\\
2B   & FT   & 0.2  & 0.8 & 0.1 & 30.7 & 0.2\\
\addlinespace
4B   & base & 7.6  & 5.0 & 0.0 & 0.0 & 44.7\\
4B   & FT   & 2.3  & 0.7 & 33.1 & 22.8 & 0.8\\
\bottomrule
\end{tabular}
\caption{Prevalence of countable failure modes per scale, variant, and exam
language, aggregated over years, retrieval conditions, and runs
(\texttt{analysis/error\_prevalence.py}).}
\label{tab:prevalence}
\end{table}

\subsection{Paired Analysis of Language Drift}
\label{app:language-drift}

Table~\ref{tab:language-drift} gives the main rates and effect sizes.
Table~\ref{tab:drift-significance} adds their exact counts and analysis
details. One cluster contains the nine Bangla outputs for an original exam
question: three retrieval conditions by three seeds. Each of 200{,}000 paired
randomizations swaps the base and fine-tuned labels for whole question
clusters. The statistic is the difference in majority-English rates among
outputs with visible non-think text, recomputed after every swap. Holm
adjustment is across the three scales.

\begin{center}
\centering
\small
\setlength{\tabcolsep}{6pt}
\begin{tabular}{lrrrr}
\toprule
\textbf{Scale} & \textbf{Base drift} & \textbf{FT drift} & $\boldsymbol{\Delta}$ \textbf{points} & $\boldsymbol{p_{\mathrm{adj}}}$ \\
\midrule
0.8B & 819/1{,}540 (53.2\%) & 12/1{,}747 (0.7\%) & -52.5 & $<.001$ \\
2B   & 764/1{,}738 (44.0\%) & 4/1{,}795 (0.2\%)  & -43.7 & $<.001$ \\
4B   & 792/1{,}771 (44.7\%) & 9/1{,}204 (0.7\%)  & -44.0 & $<.001$ \\
\bottomrule
\end{tabular}
\captionsetup{hypcap=false}
\captionof{table}{Language drift on Bangla exams. Counts are majority-English visible
outputs divided by outputs with visible non-think text. Tests are two-sided,
question-clustered paired randomization tests
(\texttt{analysis/language\_drift\_significance.py}).}
\label{tab:drift-significance}
\end{center}

Table~\ref{tab:prevalence} aggregates over retrieval conditions and years;
the leakage-delta correlation reported in Section~\ref{sec:discussion}
(Spearman $\rho=-0.47$) instead uses the per-cell fine-tuned leakage rates
of Table~\ref{tab:leakcells}, one rate per scale, exam, and retrieval
condition, paired with that cell's fine-tuned-minus-base strict delta
(\texttt{analysis/leakage\_correlation.py}).

\begin{table}[ht]
\centering
\small
\setlength{\tabcolsep}{5pt}
\begin{tabular}{ll*{4}{r}}
\toprule
& & \multicolumn{2}{c}{\textbf{2022}} & \multicolumn{2}{c}{\textbf{2023}}\\
\cmidrule(lr){3-4}\cmidrule(lr){5-6}
\textbf{Scale} & \textbf{Condition} & English & Bangla & English & Bangla\\
\midrule
0.8B & None  & 1.7 & 5.0 & 4.0 & 4.3\\
0.8B & BM25  & 5.7 & 0.7 & 4.7 & 1.3\\
0.8B & FAISS & 2.3 & 4.0 & 4.0 & 2.3\\
\addlinespace
2B   & None  & 34.3 & 0.3 & 38.3 & 0.3\\
2B   & BM25  & 26.7 & 0.0 & 43.0 & 0.0\\
2B   & FAISS & 21.7 & 0.0 & 20.0 & 0.0\\
\addlinespace
4B   & None  & 39.3 & 16.0 & 39.7 & 13.0\\
4B   & BM25  & 10.7 & 38.0 & 18.0 & 36.7\\
4B   & FAISS & 11.3 & 53.0 & 18.0 & 42.0\\
\bottomrule
\end{tabular}
\caption{Fine-tuned thinking-mode leakage rate (\%) per evaluation cell:
share of the 300 outputs (three runs of 100 questions) containing a
model-emitted \texttt{<think>} block.}
\label{tab:leakcells}
\end{table}

The stored predictions come from the original permissive extractor, so we
went back and re-checked all 21,600 of them against the letter-plus-text
contract of Section~\ref{sec:setup}
(\texttt{analysis/pair\_validate.py}). The pairing confirms 83.5\%.
Another 5.4\% parse under neither reading, 6.8\% carry a stored letter the
pairing cannot validate, and 2.0\% validate where the stored prediction is
null; the remaining 2.4\% disagree outright. We read a stratified sample
of forty disagreements by hand. Twelve favored the stored reading, ten the
pairing, seventeen were ambiguous or cut off before any verdict, and one
matched neither; exactly one of the forty came from a misread citation.
The reported tables keep the stored predictions. Re-parsing everything the
other way moves the affected cells by a few points, in directions that
leave every qualitative finding intact and, if anything, slightly enlarge
the 4B English regressions.

\section{Reproducibility Inventory}
\label{app:repro}

The complete public release, including all code, data, corpus, fine-tuning files, and per-question evaluation outputs, is available at \url{https://github.com/m-mahadi/bangladesh-legal-qa}. The project archive contains the structured law corpus, six QA splits totaling
2,165 records, two chat-format SFT files, and four evaluation files. Executed
training notebooks and trainer summary logs are preserved for all three
scales, 0.8B, 2B, and 4B; step-level metric histories are not. Executed
evaluation notebook sources and per-question outputs are preserved locally
for all three scales, 0.8B, 2B, and 4B. We recomputed the three
reported tables from those outputs, and they matched our stored
score report. The paired significance analysis is preserved as
\texttt{analysis/strict\_mcnemar.py}; its 36-cell output is
\texttt{analysis/strict\_mcnemar\_holm.csv}, and its scale-level output is
\texttt{analysis/scale\_level\_significance.csv}. The parser audit,
error-prevalence analysis, leakage-delta correlation, and language-drift
test are preserved as \texttt{analysis/pair\_validate.py},
\texttt{analysis/error\_prevalence.py},
\texttt{analysis/leakage\_correlation.py}, and
\texttt{analysis/language\_drift\_significance.py} with their CSV outputs.
Missing artifacts include the executed GPT OSS evaluation
notebook and per-item human-review records.

\section{Original Bangla Prompt Artifacts}
\label{app:prompts}

The final Bangla generation prompts survive as retained UTF-8 source files
from the original report workflow and are rendered directly from those
files, preserving the authors' wording and Bengali script. Each prompt
spans two pages so the text can be inspected at normal page size.

\clearpage
\begin{center}
\includegraphics[width=.90\textwidth,height=.82\textheight,keepaspectratio]{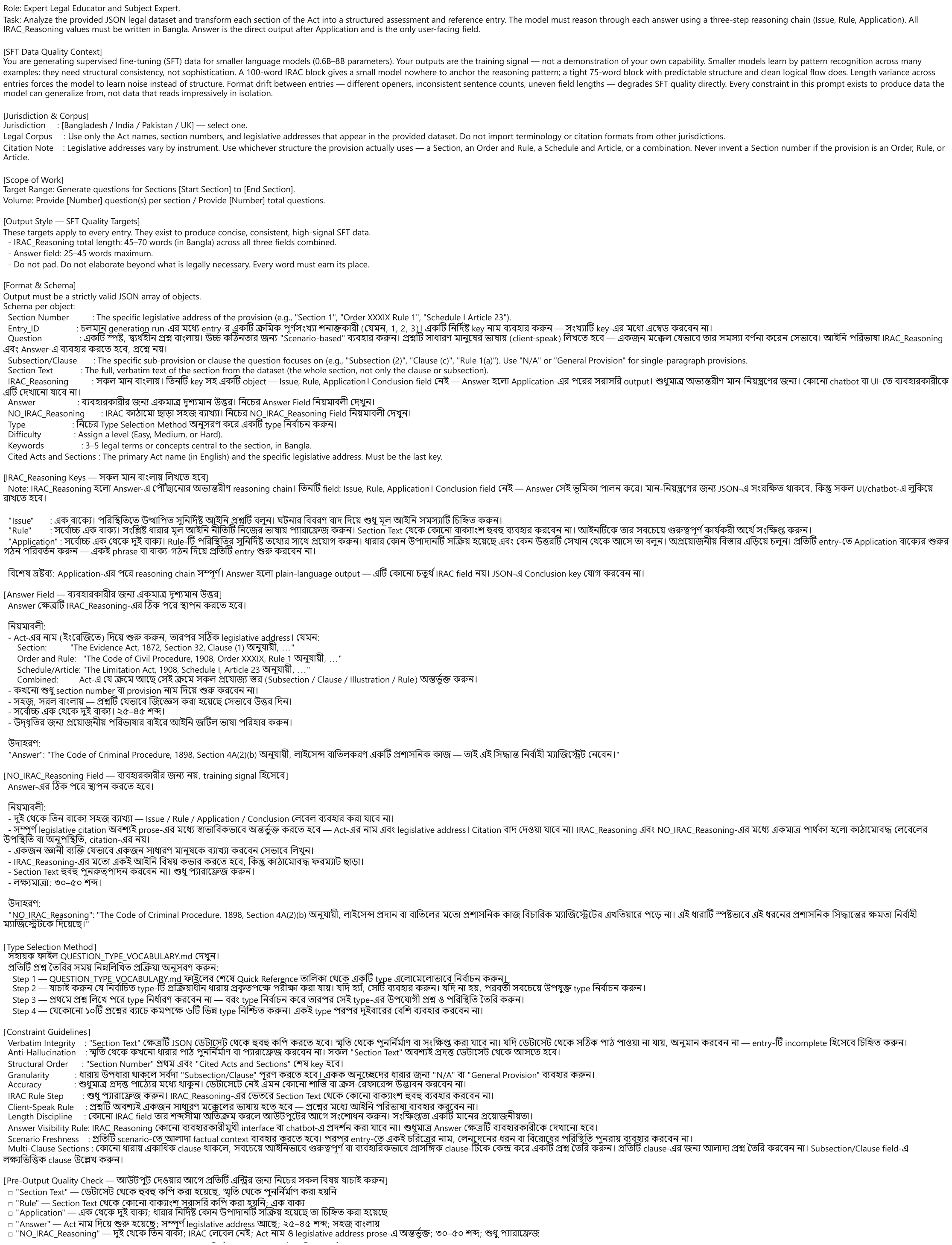}\\[4pt]
{\small Bangla single-hop generation prompt, part 1 of 2.}
\end{center}

\clearpage
\begin{center}
\includegraphics[width=.90\textwidth,height=.82\textheight,keepaspectratio]{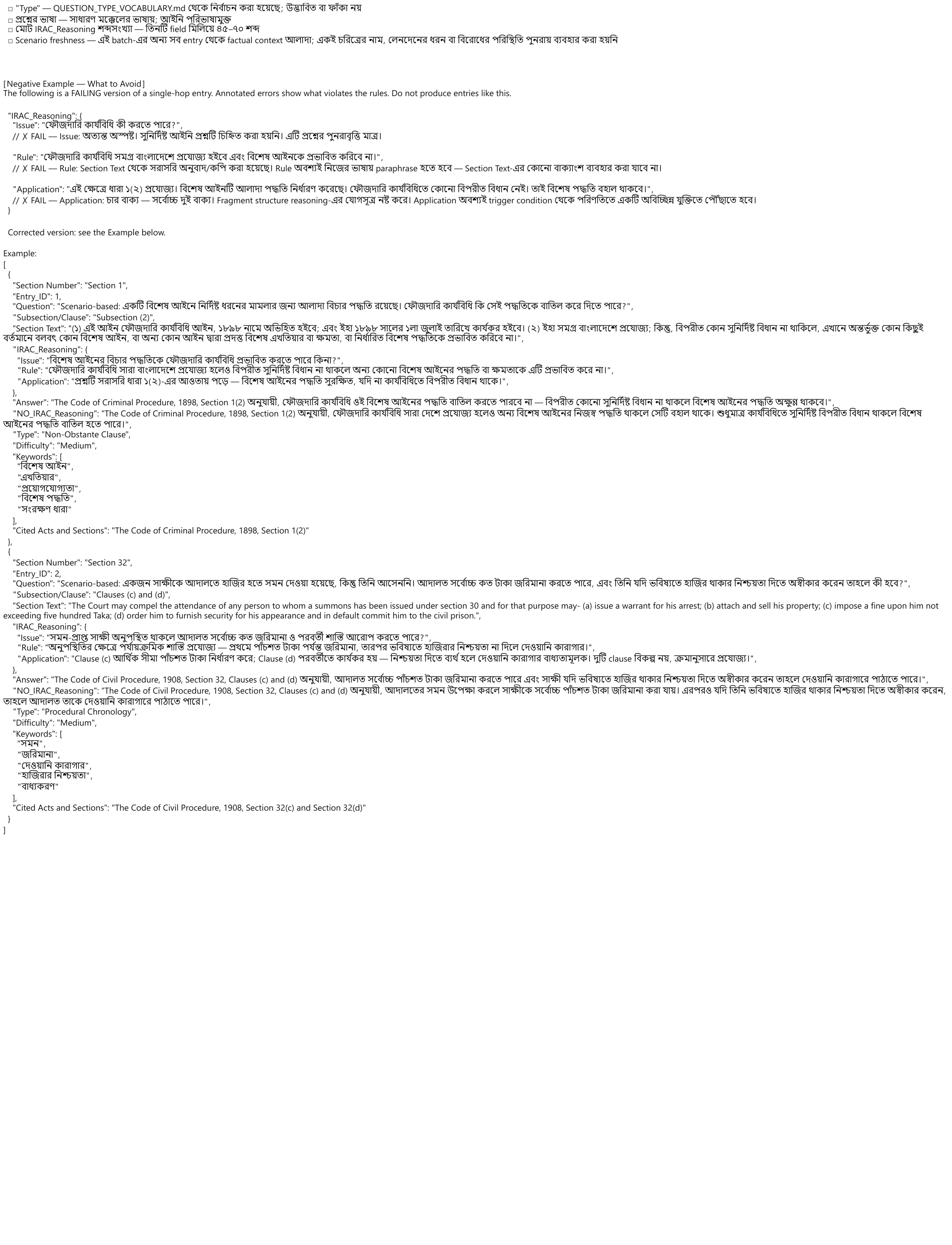}
\captionsetup{hypcap=false}
\captionof{figure}{Bangla single-hop generation prompt, part 2 of 2,
reproduced from the original report.}
\label{fig:prompt-singlehop}
\end{center}

\clearpage
\begin{center}
\includegraphics[width=.90\textwidth,height=.82\textheight,keepaspectratio]{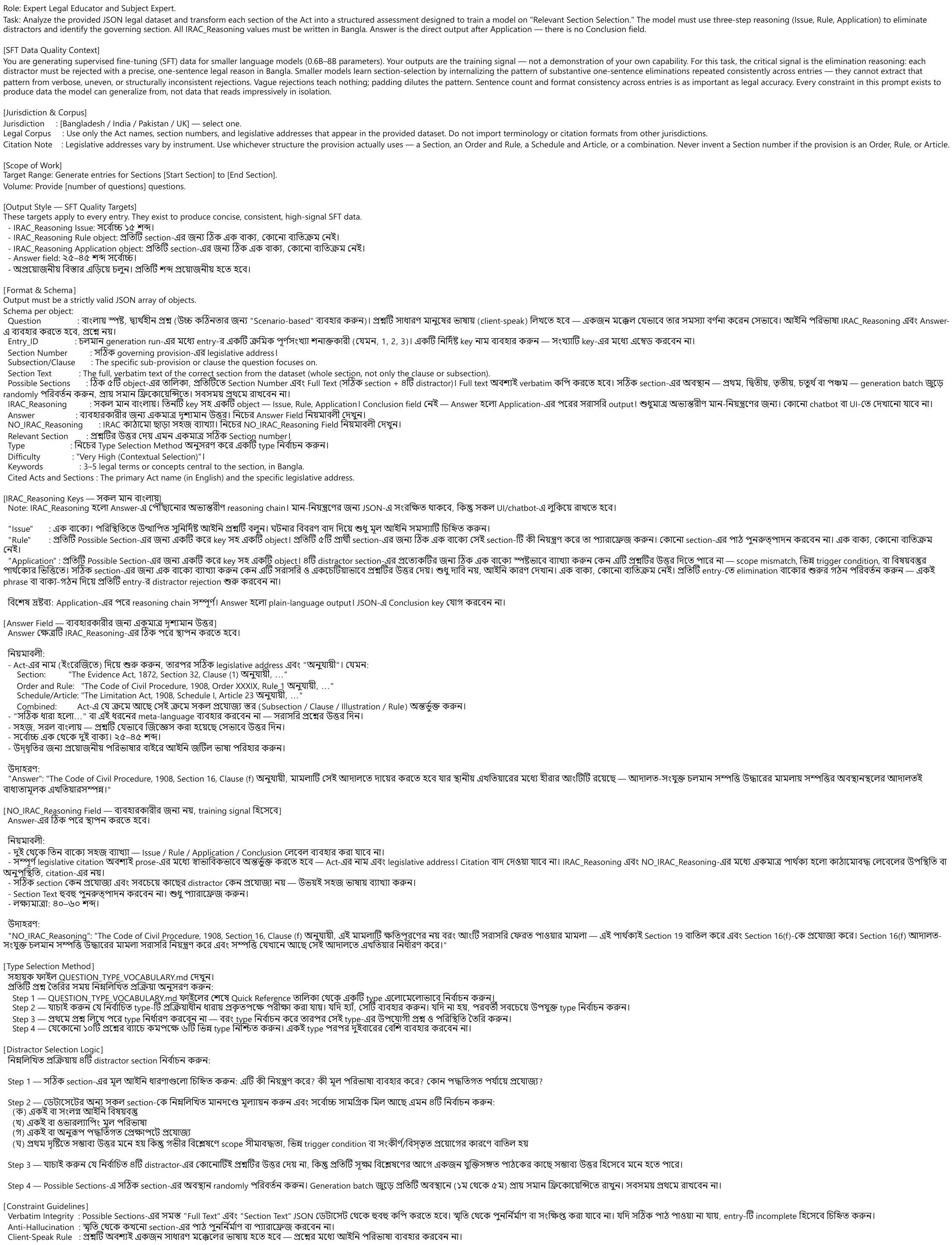}\\[4pt]
{\small Bangla advanced-selection generation prompt, part 1 of 2.}
\end{center}

\clearpage
\begin{center}
\includegraphics[width=.90\textwidth,height=.82\textheight,keepaspectratio]{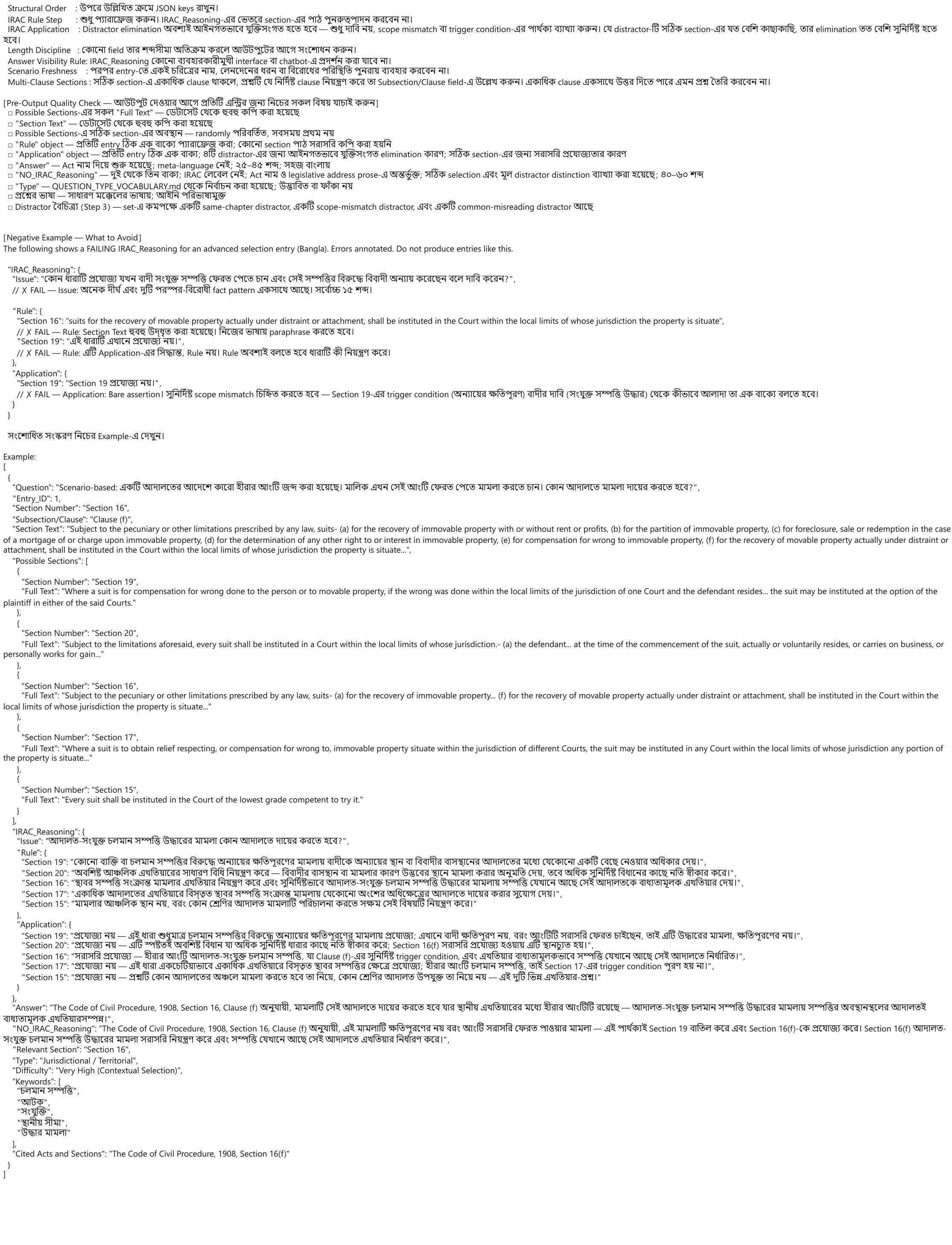}
\captionsetup{hypcap=false}
\captionof{figure}{Bangla advanced-selection generation prompt, part 2 of 2,
reproduced from the original report.}
\label{fig:prompt-advsel}
\end{center}

\clearpage
\begin{center}
\includegraphics[width=.90\textwidth,height=.82\textheight,keepaspectratio]{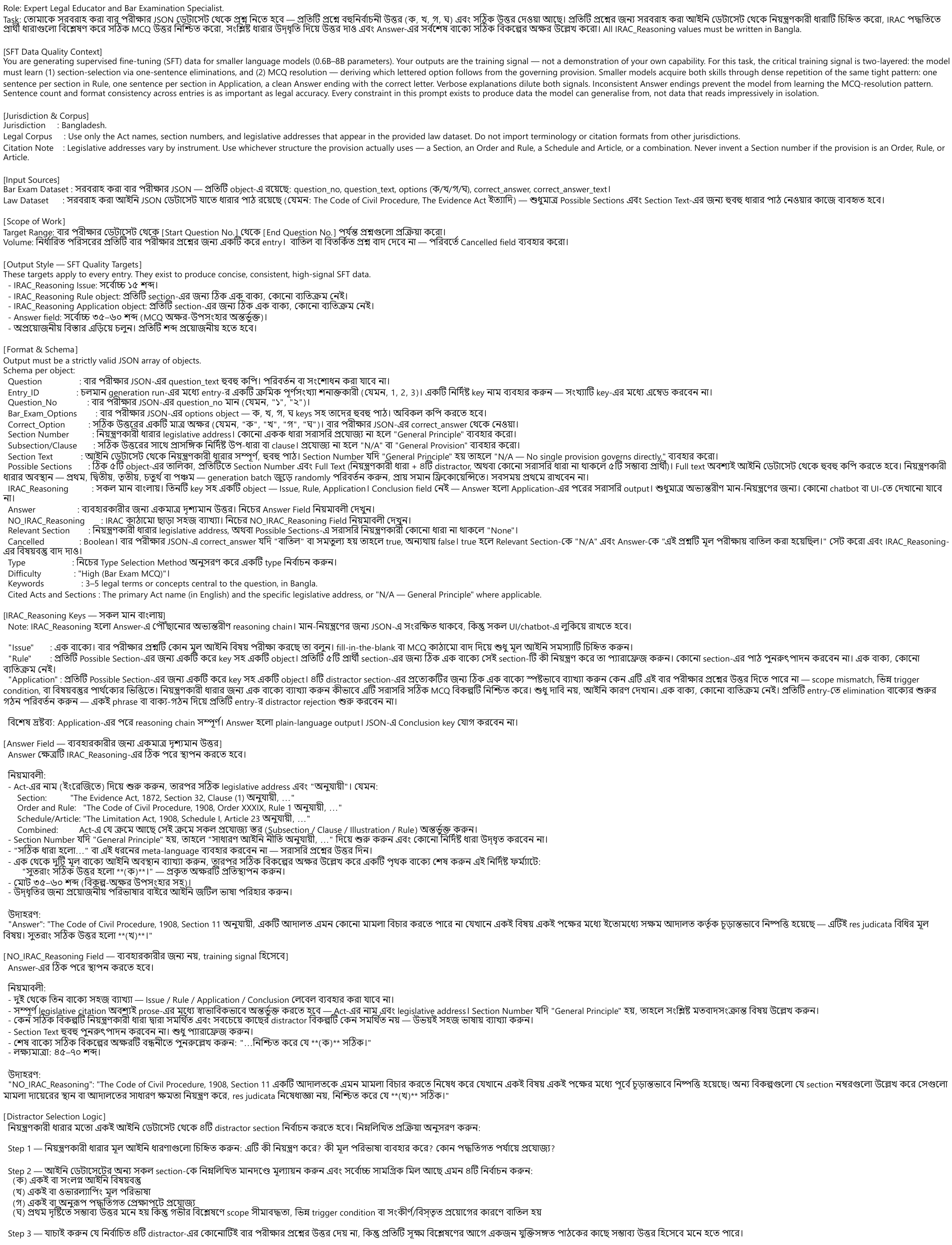}\\[4pt]
{\small Bangla bar-exam-style generation prompt, part 1 of 2.}
\end{center}

\clearpage
\begin{center}
\includegraphics[width=.90\textwidth,height=.82\textheight,keepaspectratio]{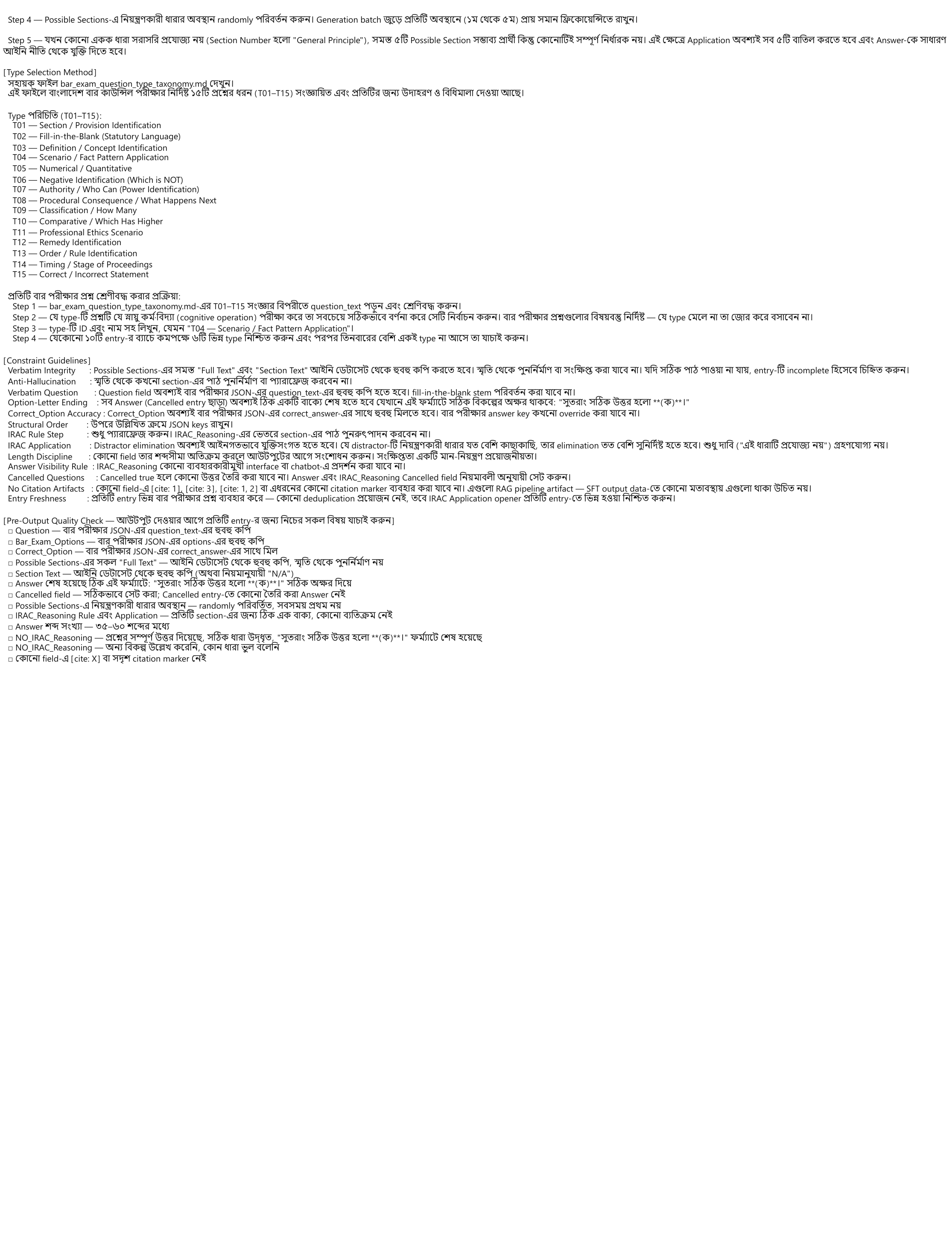}
\captionsetup{hypcap=false}
\captionof{figure}{Bangla bar-exam-style generation prompt, part 2 of 2,
reproduced from the original report.}
\label{fig:prompt-barexam}
\end{center}

\clearpage
\section{Original Error Exhibits}
\label{app:errors}

We inspected these traceable outputs without a sampling frame or coding
protocol, so they illustrate failure modes rather than estimate prevalence.
In a 2022 English item, BM25 supplied the Penal Code's spousal exception to
both 0.8B models. The base model repeated the provision but chose ``siblings''
in two of three runs; the fine-tuned model applied it and chose
``husband/wife'' in all three. A correct option still does not prove faithful
use of the statute.

A 2022 Bangla item combines several failures. The fine-tuned 2B model gave a
wrong answer with a fabricated subsection. The base model reached the correct
option with faulty reasoning in English, and the parser captured a letter
inside the fabricated citation rather than the final choice. Accuracy alone
does not measure these citation, language, reasoning, and parsing failures,
which is why the headline result remains bounded by expert review
concerns~\citep{aftahee2025}.

\begin{figure}[ht]
\centering
\includegraphics[width=.80\textwidth]{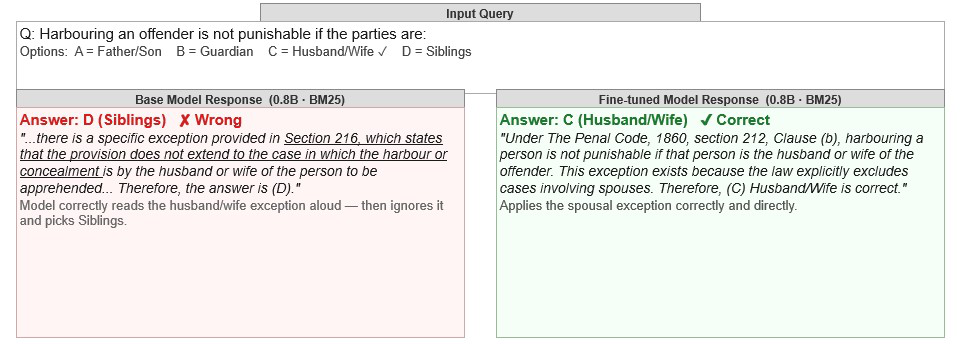}\\[1pt]
{\footnotesize (a) 0.8B, English item, BM25}\\[8pt]
\includegraphics[width=.80\textwidth]{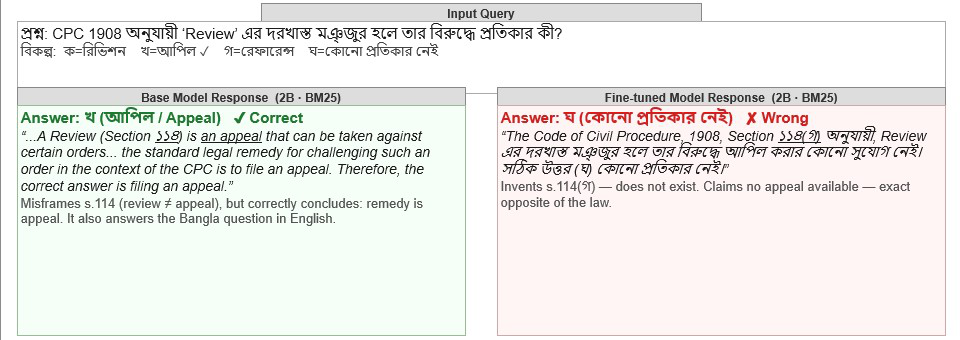}\\[1pt]
{\footnotesize (b) 2B, Bangla item, BM25}\\[8pt]
\includegraphics[width=.80\textwidth]{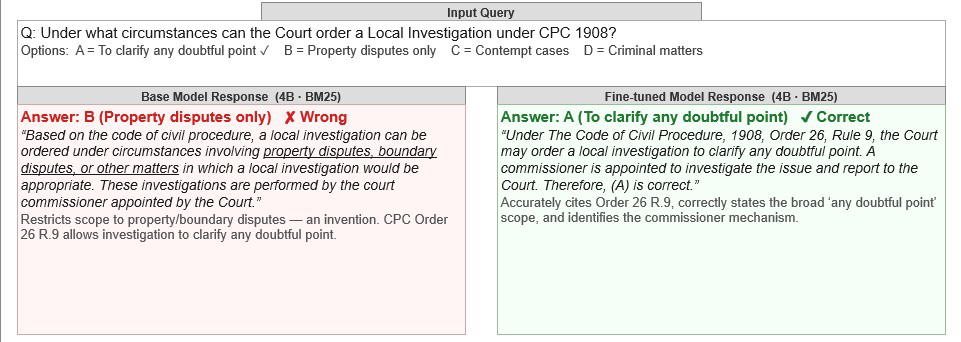}\\[1pt]
{\footnotesize (c) 4B, English item, BM25}\\[2pt]
\caption{Original error-analysis exhibits from the prior report, preserved
as images. Exhibits (a) and (b) correspond to the illustrative cases in
Section~\ref{sec:discussion}. Exhibit (c) is preserved as an artifact but
is not used for a textual claim because its source run could not be traced.}
\label{fig:errors}
\end{figure}